\documentclass{article}

\usepackage[nonatbib,final]{neurips_2019}

\usepackage[utf8]{inputenc} 
\usepackage[T1]{fontenc}    
\usepackage{hyperref}       
\usepackage{url}            
\usepackage{booktabs}       
\usepackage{amsfonts}       
\usepackage{nicefrac}       
\usepackage{microtype}      

\usepackage[numbers]{natbib} 
\usepackage{wrapfig}
\usepackage[pdftex]{graphicx}
\usepackage{algorithm,algorithmic}
\usepackage{amsmath,amsthm,amsfonts}
\usepackage{color,enumitem}
\graphicspath{{figures/}}
\DeclareMathOperator*{\argmax}{arg\,max}

\usepackage{verbatim}

\title{ZPD Teaching Strategies for Deep Reinforcement Learning from Demonstrations}

\author{%
 Daniel Seita, David Chan, Roshan Rao, Chen Tang, Mandi Zhao, John Canny\\
  University of California, Berkeley \\
  \texttt{\{seita,davidchan,roshan\_rao,chen.tang,mandi.zhao,canny\}@berkeley.edu} \\
}

\begin{document}

\maketitle

\begin{abstract}
Learning from demonstrations is a popular tool for accelerating and reducing the exploration requirements of reinforcement learning. When providing expert demonstrations to human students, we know that the demonstrations must fall within a particular range of difficulties called the ``Zone of Proximal Development (ZPD)''~\cite{chaiklin,vygotski}. If they are too easy the student learns nothing, but if they are too difficult the student is unable to follow along. This raises the question: Given a set of potential demonstrators, which among them is best suited for teaching any particular learner? Prior work, such as the popular Deep Q-learning from Demonstrations (DQfD) algorithm has generally focused on single demonstrators. In this work we consider the problem of choosing among multiple demonstrators of varying skill levels. Our results align with intuition from human learners: it is not always the best policy to draw demonstrations from the best performing demonstrator (in terms of reward). We show that careful selection of teaching strategies can result in sample efficiency gains in the learner's environment across nine Atari games. 
\end{abstract}

\section{Introduction}
When a child follows an adult teaching by example, we want the child to gradually develop skills without assistance. Intuitively, an effective teaching method might provide examples that are within a ``comfort zone'' for the student --- just difficult enough to learn from, but not so difficult that the student cannot benefit from the teaching. This intuition is formalized in ``Zone of Proximal Development (ZPD)'' theory, an epistemological theory built on a human cognitive model first proposed by social psychologist Lev Vygotsky~\cite{chaiklin,vygotski}. According to ZPD theory, learning examples are placed on a spectrum with problems that a learner can solve on their own on one side, and problems a learner cannot solve on the other. According to Vygotsky, when presenting a student with examples, those examples must be ``in the Zone of Proximal Development,'' a set of problems which are difficult for the learner to solve alone, but can be solved with guidance.

In the proposed work, we investigate the application of ZPD in deep reinforcement learning contexts where a learner agent has access to several teacher agents or demonstrators. The primary question we explore is the following: which among a given set of teachers should the agent ``select'' to best improve learning?  While there has been substantial work on reinforcement learning with demonstrators (e.g., DQfD~\cite{dqfd} and DDPGfD~\cite{leveraging_demos_2017}), to our knowledge there is no work investigating which among a set of teachers should be selected for teaching. We focus on off-policy Deep Q-learning algorithms with experience replay~\cite{Lin1992} to act both as teacher agents, and as a student agent. We evaluate on nine Atari 2600 benchmarks~\citep{bellemare13arcade}. Our contributions include:

\begin{enumerate}[noitemsep]
\item A ZPD-based algorithm that allows for dynamic selection of teachers for a given student agent.
\item Results showing that choosing the appropriate teaching strategy, which may not always be a single teacher, improves sample efficiency and may improve final reward.
\end{enumerate}


\section{Related Work}\label{sec:references}

We focus on applying ZPD concepts in off-policy deep reinforcement learning algorithms with experience replay, which allows sharing of samples between learners and teachers. Prominent examples of (deep) off-policy algorithms include the seminal Deep Q-Network (DQN)~\cite{mnih-dqn-2015} and its notable extensions, such as Double DQN~\cite{van2016deep} which we employ in this work. Other improvements are accounted for in algorithms such as Rainbow~\cite{rainbow} which combined multiple algorithmic components and showed complementary benefits. For simplicity, we use Double DQN (DDQN) as the base algorithm, but the algorithm we propose is broadly applicable to any off-policy, experience replay based agent.

Despite various advances in deep reinforcement learning, sample efficiency and exploration remain known challenges. A common ingredient to improve sample efficiency is using a demonstrator (equivalently in this work, a teacher). DAgger~\cite{ross2010reduction} is a popular approach, but requires a teacher to be continually present to label student-generated states with actions. Arguably a more practical alternative is to assume a fixed batch of data from (potentially multiple) demonstrators, without requiring any further teacher interaction, and to see how much an agent can learn from the data. If the agent does not do any environment interaction, the setting is known as \emph{batch reinforcement learning}~\cite{off-policy-deeprl-2019,offpolicy_atari_2019,stabilizing-qlearning_2019-neurips}.

We similarly focus on off-policy learning with demonstrations, but allow the agent to take environment steps in addition to using demonstrator samples for learning. A prominent example of this for discrete control is Deep Q-learning from Demonstrations (DQfD)~\cite{dqfd}, the most relevant prior work, which seeded a learner agent with a small batch of human demonstrator data and showed performance gains on Atari games. Follow-up work, including Ape-X DQfD~\cite{ape-x_dqfd,distributed_per} and R2D3~\cite{r2d3} scaled the idea to handle a larger stream of samples and a wider variety of environments. In continuous control several algorithms have built upon the Deep Deterministic Policy Gradients (DDPG)~\cite{lillicrap2015continuous} algorithm such as in~\cite{leveraging_demos_2017,overcoming,socket_insertion_2018} to effectively leverage demonstrations for sparse reward tasks. Schmitt~et~al.~\cite{kickstarting2018} apply a similar idea in the actor-critic setting by adding an auxiliary cross-entropy loss function from~\cite{pd_2016} to encourage learner policies to be closer to a teacher's policy, and additionally consider an ensemble of teachers specialized to different tasks. None of these prior works, though, primarily focuses on determining which among a \emph{set} of teachers we should select for using samples.

The problem setting we consider can more generally be viewed as that of using a neural network teacher to accelerate the training of a neural network student. One of the most straightforward methods is to train a student model so that its final layer matches that of the teacher network. This is known as \emph{distillation} and has been applied in classification~\cite{distillation_2014,bann} and reinforcement learning~\cite{pd_2016} to decrease the size of neural networks and to improve task generalization. The method we propose is distinct but complementary to distillation. We focus on teaching via samples in a shared replay environment rather than matching final neural network layers, but the method of selecting appropriate teachers could be used for distillation based approaches, forming a curriculum~\cite{bengio_2009} over teachers.

\section{Methods}\label{sec:methods}

In a setup mirroring DQfD~\cite{dqfd}, we train student agents using DDQN~\cite{van2016deep} as the base algorithm, where training is based on a mixture of self-generated on-policy data and off-policy data from a teacher. Unlike in DQfD, we update the distribution of teacher data according to a ``teaching strategy'' $f_{\rm select}$ which draws experience from different teachers depending on the performance of the student. This teacher experience is blended in a mini-batch with student experience according to a second function $f_{\rm blend}$. This differs from the classical literature in that the teacher experience now comes from a dynamic distribution instead of a fixed distribution drawn from the best performing teacher agent. An outline of the method is given in Algorithm~\ref{alg:general}.

\newcommand{\D}{\mathcal{D}}

\begin{algorithm}[tb]
\caption{The Student-Teacher Framework}
\label{alg:general}
\begin{algorithmic}[1]
\STATE {\bfseries Input:} A set of data from teacher snapshots, $\{\D_{T_1}, \ldots, \D_{T_N}\}$ corresponding to the teacher's training run, student policy $\pi_\theta$, student replay buffer $\D_{\rm S}$, gradient update frequency $u$.
\STATE $T_{\rm ZPD} := f_{\rm select}(\pi_\theta)$
\FOR{steps $t \in \{1, 2, \ldots\}$}
    \STATE Take an environment step using $\pi_\theta$, store resulting sample in $\mathcal{D}_{\rm S}$
    \IF{$t \mod u = 0$}
        \STATE $\D_{\rm minibatch} := f_{\rm blend}(\D_{T_{\rm ZPD}}, \D_{\rm S})$
        \STATE Compute gradient update with $\D_{\rm minibatch}$
        \STATE $T_{\rm ZPD} := f_{\rm select}(\pi_\theta)$
    \ENDIF
\ENDFOR
\end{algorithmic}
\end{algorithm}

\subsection{Background}

We use the standard Markov Decision Process (MDP) formulation for reinforcement learning~\cite{Sutton_2018}. An MDP consists of a five-tuple $(\mathcal{S}, \mathcal{A}, P, R, \gamma)$ where at each time step $t$, the agent is at a state $s_t \in \mathcal{S}$ from the set of possible states. For our benchmark Atari environments, the states are of dimensions $\mathbb{R}^{84\times 84\times 4}$, and represent gray-scale environment images, consisting of four frames. The agent takes a (discrete) action $a_t \in \mathcal{A}$ from the set of possible actions, which in the Atari environments are game dependent. The environment dynamics map the state-action pair into a successor state $s_{t+1} \sim P(\cdot | s_t, a_t)$, and the agent receives a scalar reward signal $r_t = R(s_t,a_t)$. The objective is to find a policy $\pi : \mathcal{S} \to \mathcal{A}$ that maximizes the expected discounted return $\mathbb{E}[\sum_{t=0}^\infty \gamma^t r_t]$ with $\gamma \in (0,1]$.

\subsection{The Student and Teacher Agents}\label{ssec:student_and_teacher}

We consider the RL framework with \emph{student}, $S$, and \emph{teacher}, $T$, agents and denote policies as $\pi_\psi$ with parameters $\psi$, which could represent either the student or the teacher. In our experiments we use the same neural network architecture for $S$ and $T$, but this is not an assumption of the method, which treats the teacher as a black-box demonstrator. 

In this paper we use a set of snapshots of a standard DDQN~\cite{van2016deep} agent pre-trained on the benchmark environment to serve as teachers. The teacher training uses the following loss function $J_T$ for a given sample $(s,a,r,s')$ to adjust its Q-network $Q$:
\begin{equation}\label{eq:ddqn}
J_T(Q) = \Big(
R(s,a) + \gamma Q(s',a_{\rm max}'; \psi') - Q(s,a,; \psi')
\Big)^2
\end{equation}
where $a_{\rm max}' = \argmax_{a\in \mathcal{A}} Q(s',a; \psi)$, and $\psi$ and $\psi'$ are parameters of the current and target networks, respectively. 

In the proposed algorithm, $S$ can draw samples from both self-generated environment data, and from the teacher $T$. We train the student with a modified DDQN loss to account for the off-policy teacher samples which are injected into the student's replay buffer. The modified loss is the sum:
\begin{equation}\label{eq:student}
    J_S(Q) = J_T(Q) + \lambda_E J_E(Q) + \lambda_2 J_{L_2}(Q)
\end{equation}
where $J_E(Q)$ is the additional large margin classification loss \cite{piot2014boosted}, as used by DQfD~\cite{dqfd}:
\begin{equation}
    J_E(Q) = \max_{a \in \mathcal{A}}\Big[ Q(s,a)+\ell(a_E,a)\Big] - Q(s,a_E)
\end{equation}
and $J_{L_2}(Q)$ is an $L_2$ weight regularization of the Q-network. The $\lambda_E$ and $\lambda_2$ are hyper-parameters controlling a trade-off between these loss values.

While we have designed the DDQN agents to closely model the original DQfD paper~\cite{dqfd}, our implementation differs in two major ways:
\begin{itemize}
    \item We do not use the pre-training phase from DQfD in order to improve the generality of our results, and to explore training agents \textit{de novo}.
    \item We do not use prioritization of the replay buffer~\cite{schaul2015prioritized} in any form in order to closely examine the effect of having a fixed minibatch sampling ratio (discussed in Section~\ref{sec:minibatch}) on the performance of the model.
\end{itemize}

While it is possible to use an ensemble of fully trained teacher agents as teachers, we instead draw our teachers from a set of snapshots generated during the training of a single teacher DDQN network $T$. During the teacher training, we store snapshots of its parameters at regular time intervals with respect to environment steps. This results in a set of $N$ saved snapshots $\{T_1, \ldots, T_N \}$ from the training history of $T$, each corresponding to a different policy. For the $k$-th teacher policy $\pi_{T_k}$ from the training history, we denote a dataset of $m$ states (induced by running the policy $\pi_{T_k}$ in the environment) as 
\begin{equation}
\mathcal{D}_{T_k} = \{s_0^k,s_1^k,\ldots,s_m^{k}\} \sim \pi_{T_k}.
\end{equation}
To make results more reproducible and to facilitate comparisons among different teaching methods, we take $\mathcal{D}_{T_k}$ to be a set of samples that the teacher agent experienced in a window around the time that the snapshot was saved. 

\subsection{Student/Teacher Matching with $f_{\rm select}$}\label{sec:select}

The ZPD teacher selection function $f_{\rm select}$ is a critical feature of Algorithm~\ref{alg:general}. This function determines the distribution across teachers from which we sample experience. In our experiments, we draw only from a delta distribution across the teachers, and refer to the selected teacher as the \emph{ZPD} teacher $T_{\rm ZPD}$. We leave exploring prioritized sampling across multiple teachers to future work.

We propose defining $f_{\rm select}$ as a ``fixed snapshots ahead'' strategy, with an integer hyperparameter $k$. For any student $S$, that student is matched to a snapshot $T_{\rm match}$, the teacher with the reward closest to the one that the student achieves. We refer to the snapshot $T_{\rm match}$ as the ``matched'' teacher snapshot. The $f_{\rm select}$ function sets the ZPD teacher (the snapshot from which experience is drawn) to be snapshot $T_{{\rm match} + k}$, capped at the last snapshot if it exceeds the total number of snapshots. Intuitively, the currently matched snapshot tells us the teacher snapshot that is roughly on par with the student, and so the next ZPD snapshot may be some fixed interval ahead, with higher rewards indicating that we are using teachers that are more skilled at a given task. Another interpretation of this matching function is as a curriculum~\cite{bengio_2009} where the learner initially starts with a teacher that provides ``easier'' samples, before moving on to harder teachers.

While it is possible that updating the ZPD teacher at each time-step would provide the best performance, for efficiency, we update the ZPD teacher infrequently. A call to $f_{\rm select}$ triggers when the student's average reward (over a window of 100 episodes) exceeds that of the matched teacher's reward, the latter of which is the average over a window of 100 episodes the teacher experienced at the time the snapshot was saved.

In this work, we examine fixed snapshot ahead strategies for $k \in \{0,2,5,10\}$ and call these ``$k$ ahead'' methods. We additionally test with two baselines: a ``best ahead'' where the ZPD snapshot is fixed throughout student training to be the teacher snapshot with highest reward, and a ``random ahead'' where the next ZPD snapshot is chosen at random among the set of all teacher snapshots with higher reward than the student.

\subsection{Minibatch Sampling with $f_{\rm blend}$}\label{sec:minibatch}

Once a ZPD teacher, $T_{\rm ZPD}$, is selected, we can treat the corresponding dataset $\mathcal{D}_{T_{\rm ZPD}}$ of samples as a distinct ``ZPD teacher'' replay buffer. We then sample from both $\mathcal{D}_{\rm S}$ and $\mathcal{D}_{T_{\rm ZPD}}$ in each minibatch update of the DQfD algorithm. When sampling, however, we need to decide how many samples to draw from the teacher's replay buffer (off-policy examples) and how many to draw from the replay buffer of the student (on-policy examples). This trade-off is made according to the strategy $f_{\rm blend}$, a binomial distribution representing the probability that a sample in the minibatch is drawn from a particular replay buffer.

In this work, we use a fixed ratio of student-teacher samples per mini-batch. While prior work has tested using a ratio of 25\% teacher samples in a minibatch as in Ape-X DQfD~\cite{ape-x_dqfd} or even smaller ratios of 1/256 as in R2D3~\cite{r2d3}, we use a 50\% ratio to test the effect of using more off-policy teacher data per minibatch. These ratios are held constant throughout student training, up to the point when the student surpasses every teacher snapshot. When this occurs, we anneal $f_{\rm blend}$, in a linear decay towards no teacher samples. At this point, the student has little to learn from the teachers.

\section{Experimental Setup}\label{sec:setup}

\begin{figure}[t]
\centering
\includegraphics[width=1.00\columnwidth]{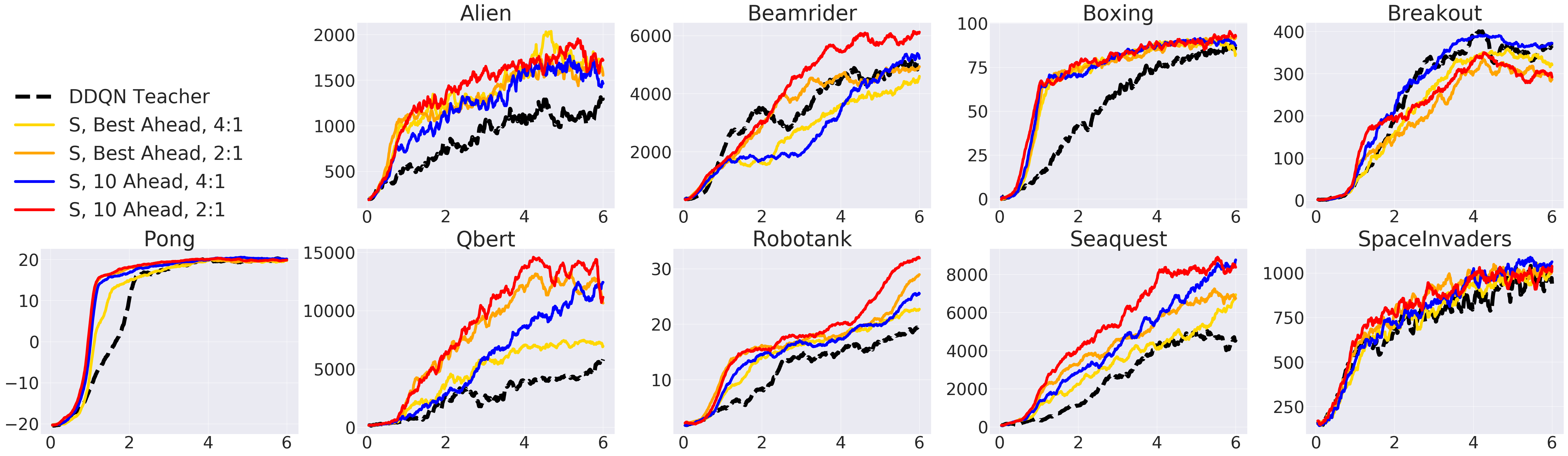}
\caption{
Learning curves for all nine games tested, comparing: 10 ahead, best ahead, and the original DDQN teacher. For 10 ahead and best ahead, we test with both a 4:1 and a 2:1 ratio of steps:updates, as discussed in Section~\ref{sec:setup}. Learning curves are over two random seeds. Results suggest that the 2:1 throughput case is at or exceeds the level of the 4:1 case across games.
}
\label{fig:throughput_1}
\end{figure}

We evaluate our methods using the Atari~\cite{bellemare13arcade} benchmarks for the following nine games: Alien, BeamRider, Boxing, Breakout, Pong, Qbert, Robotank, Seaquest, and SpaceInvaders. These include several widely-used games (Pong and Breakout in particular) and others for which DDQN can get reasonable performance. We run into a limitation for complex games with sparse rewards, such as Montezuma's Revenge and Private Eye. For those games, training a teacher agent via deep reinforcement learning is famously difficult, and successful learning normally requires some combination of human demonstrators, the underlying game state, or advanced exploration strategies with massively parallel data collection~\cite{montezuma_revenge,rnd_2019}. Since we compare performance of students versus teachers, we wish to have teachers follow the same base algorithm of DDQN, so we did not test on games for which DDQN performs extremely poorly.

For all teacher DDQN training runs, we saved 23 snapshots, spaced out over 250,000 environment training steps out of the 6 million total steps taken, where each environment step is exactly 4 frames, following OpenAI gym~\cite{openaigym} conventions. The first teacher snapshot was saved at 400,000 steps, the second at 750,000 steps, and continued until the 23rd at 5,900,000 steps. We use 30 random no-op actions to begin each environment. For each teacher selection method, we benchmark with two random seeds and average the results.

\begin{table}[t]
\caption{
Hyperparameters used when training teachers and then students. In addition to values listed here, both the student and teachers used an Adam~\cite{adam} learning rate of $10^{-4}$, batch size 32, target network sync-ing period of 10,000 steps, and discount factor $\gamma = 0.99$.
}\smallskip
\centering
\small
\begin{tabular}{l r r}
\textbf{Hyperparameter} & \textbf{Teacher} & \textbf{Student} \\  \hline
Environment steps & 6,000,000 & 6,000,000 \\
Replay buffer size & 1,000,000 &  250,000 \\
Steps between gradient updates & 4  & 2 \\
Student-teacher minibatch ratio & N/A & 50-50  \\
$L_2$ regularization weight $\lambda_{L_2}$ & N/A & $10^{-5}$ \\
Supervised loss weight $\lambda_{E}$ & N/A & $10^{-1}$ or $10^{-2}$ \\
Supervised loss margin $\ell$  & N/A & 0.8 \\
Size of ZPD teacher buffer $\mathcal{D}_{T_{\rm ZPD}}$ & N/A & $\approx 250000$ \\
\end{tabular}
\label{tab:rl-hyperparams}
\vspace{-2em}
\end{table}

We test the following six $f_{\rm select}$ methods: 0 ahead, 2 ahead, 5 ahead, 10 ahead, best ahead, and random ahead (discussed in Section \ref{sec:select}). For the DQfD lambda hyperparameter $\lambda_E$, which controls the strength of the imitation loss, we tuned values in $\{0, 0.001, 0.01, 0.1, 1.0\}$ for all nine games with the 0.8 large margin loss from~\cite{dqfd}. In contrast to their work, we found that $\lambda_E = 1.0$ was too large and resulted in deficient student performance, perhaps because we use relatively more off-policy data. Based on initial tuning, six games worked best with $\lambda = 0.1$ and the other three (BeamRider, Pong, Robotank) worked best with $\lambda=0.01$. Our main hyperparameters are shown in Table~\ref{tab:rl-hyperparams}, where we borrow some relevant hyperparameters from DQfD~\cite{dqfd} to minimize tuning, and reduce the size of the experience replay buffer~\cite{deeper_look}. 

We also tuned the number of environment steps taken by the student in between gradient updates. The standard in DQN is to use 4 environment steps for every gradient update. With 50\% of the minibatch coming from demonstrator samples, doing a 4:1 step:update ratio means that each new student environment step sample is consumed, in expectation, half as many times as in the no-teacher case. A 2:1 ratio allows the student to use environment samples at the same rate, and to consume teacher samples at twice the rate. We thus tested both 4:1 and 2:1 ratios across all games and teaching methods. Figure~\ref{fig:throughput_1} shows a representative set of results, suggesting that students trained with a 2:1 ratio outperform those with a 4:1 ratio. Thus, for all results we use a 2:1 ratio, and leave examining the step-to-update ratio for ``student throughput'' to future work.


\section{Results}\label{sec:results}

%
%
%
%
%

\begin{figure}[t]
\centering
\includegraphics[width=1.00\columnwidth]{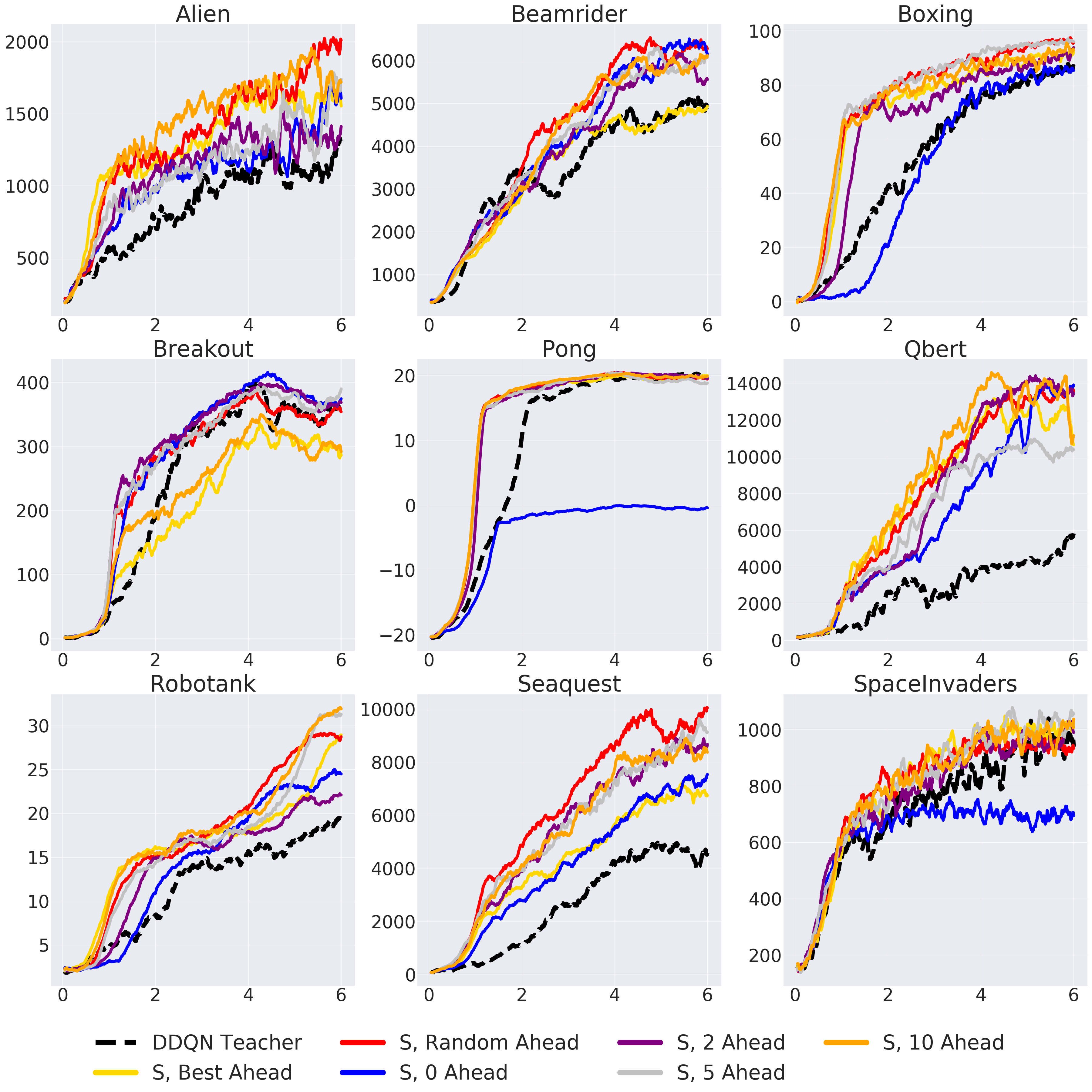}
\caption{
Results for $f_{\rm select}$ methods as described in Section~\ref{sec:select} on the nine Atari games we use. Learning curves represent the average reward over a window of 100 past episodes as a function of environment steps (six million total) as suggested in~\cite{revisiting_ale}. Curves are then re-averaged over two random seeds per $f_{\rm select}$ function. Each seed for a game is run with the \emph{same} teacher and the \emph{same} set of datasets generated from that teacher to reduce the variability in results. The black dashed lines represent the performance of the DDQN teacher agent, run only once per game. Results suggest that using the best ahead matching, indicated in bright gold above, is not the clear best choice for any of the games tested. 
}
\label{fig:example_2}
\vspace{-2em}
\end{figure}

Figure~\ref{fig:example_2} plots the learning curves, averaged over two random seeds, for all nine games and all teaching strategies tested. Traditional methods such as DQfD rely on the best ahead style of teaching --- they choose a constant teacher achieving the highest final reward. We can see that by dynamically altering the teaching strategy, in almost all cases the best ahead strategy is sub-optimal with respect to both final reward and sample efficiency. Table~\ref{tab:performance} summarizes this observation.

The learning curves and tables suggest that the strongest teaching methods are the random ahead, 5 ahead, and 10 ahead methods. This result not only supports our overall hypothesis that a dynamic ZPD style strategy can improve performance, but also indicates that it may not be necessary to use the very highest-rewarding expert available in all circumstances. One possible explanation for these results is that with high-quality samples from the teacher with highest reward, running a DQfD-style algorithm may overfit to that teacher's samples (as overfitting has been shown in deep Q-learning~\cite{bottlenecks_2019}). Also intriguing is the performance of the random ahead method. The diversity of samples introduced by the method could aid in avoiding overfitting to a given teacher or a given skill range. We believe that it is interesting future work to further explore these random teaching methods, as additional exploration could yield insights into the learning process of Deep Q Networks.

Another clear result is that, as expected, the best teaching methods for a particular game outperform the teacher itself. This is despite the 50-50 minibatch ratio, which uses a relatively large amount of off-policy data as compared to DQfD~\cite{dqfd}, Ape-X DQfD~\cite{ape-x_dqfd}, and especially R2D3~\cite{r2d3}, which complicates learning as Q-learning algorithms normally need significant on-policy experience~\cite{off-policy-deeprl-2019}.

\begin{table}[H]
\caption{
A ranking of the various methods across the nine games, combining all trials from Figure~\ref{fig:example_2}. We indicate the number of times the method ``won'', i.e., had highest reward (over two random seeds) for a given game.
}\smallskip
\small
\centering
\begin{tabular}{l | r  r  r  r  r  r}
Method   & BA & RA & 0A & 2A & 5A & 10A \\ \hline
Average  & 0 & 2 & 0 & 0 & 3 & 4 \\
Last 100 & 1 & 3 & 1 & 0 & 3 & 1 \\
\end{tabular}
\label{tab:performance}
\vspace{-2em}
\end{table}

\subsection{Per-Game Analysis}

\begin{figure}[t]
\centering
\includegraphics[width=1.00\columnwidth]{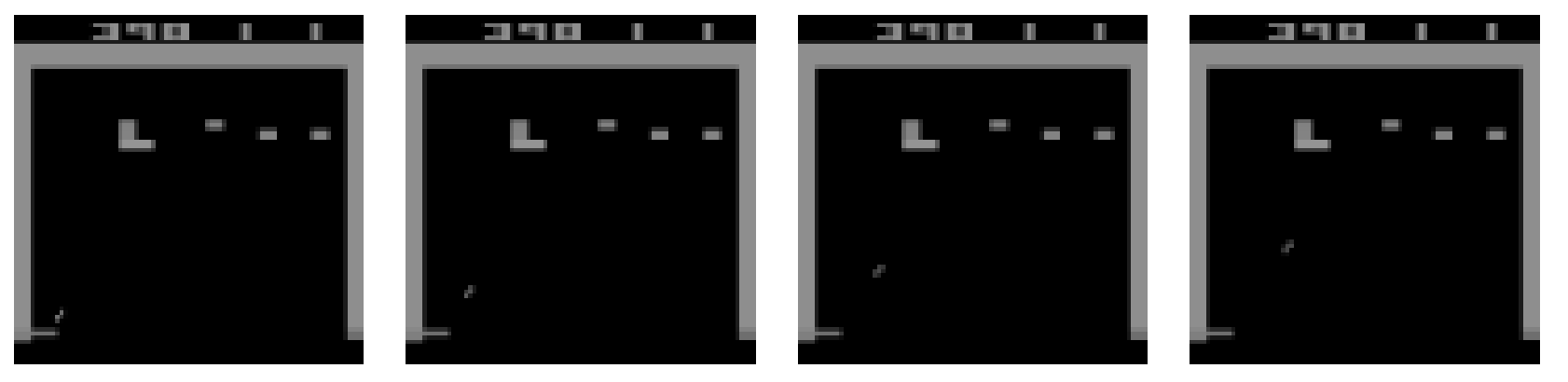} 
\caption{A representative example of a set of Breakout frames that an expert agent frequently encounters. Here, the agent moves back and forth repeatedly in the same pattern and the ball does not hit any remaining bricks. The game only exits this ``deadlock'' due to either the 1\% chance of a random action from the epsilon greedy policy or the 10K episode step limit.}
\label{fig:specific-game}
\end{figure}

\begin{figure}[t]
\centering
\includegraphics[width=1.00\columnwidth]{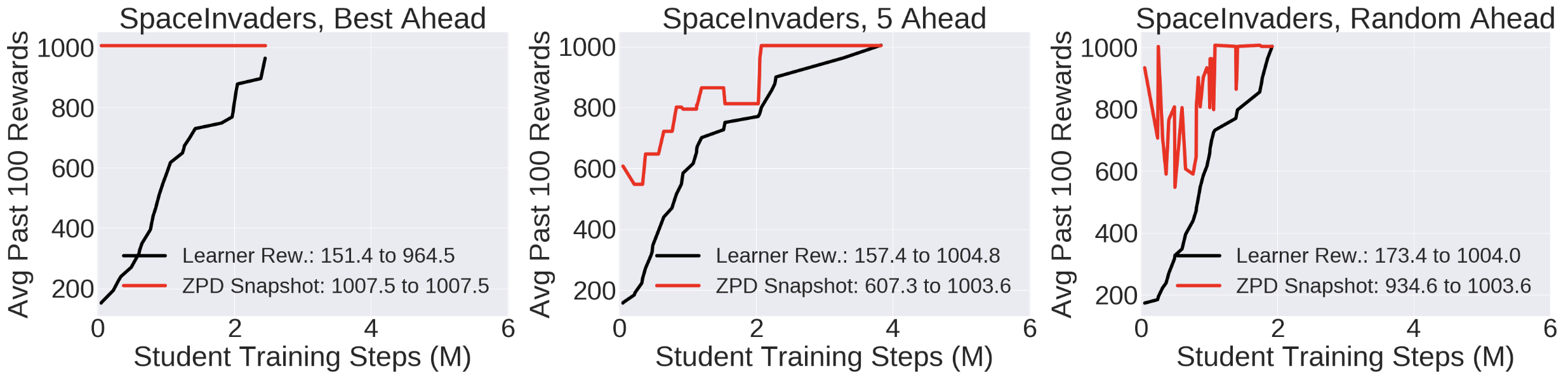} 
\caption{
Examples of learner reward curves and corresponding ZPD reward values for Space Invaders for $f_{\rm select}$ functions: best ahead (left), 5 ahead (middle), and random ahead (right). The plots only show the curves for the period when the learner has lower reward than its ZPD teacher. At an environment step, the reward of the learner and the reward of its ZPD teacher is plotted, showing that the best ahead method always uses a ZPD teacher with the highest possible reward, while the random method uses ZPD teachers with varying reward levels (that exceed the reward of the learner).
}
\label{fig:zpd}
\end{figure}

The results indicate that the best strategy for making use of teacher samples appears to be game-dependent. From Table~\ref{tab:performance}, the best ahead method is only highest in one case, Pong, when considering the average of the last 100 episodes. The learning curves indicate Pong is unique: all teaching methods, with the exception of 0 ahead, have near identical performance. This phenomenon is surprising when considering that, for example, the 2 ahead method will initially query samples from a ZPD teacher that attains roughly -17 reward. Episodic rewards for Pong range from -21 to 21, so this suggests that using samples from teachers with low game rewards can nonetheless be beneficial.

Breakout is another interesting case study. The learning curves suggest that the best ahead and the 10 ahead strategies are the two weakest strategies for the learner. One hypothesis for why Breakout may be better suited to a shorter ZPD ``advancement interval'' is because more advanced agents are frequently in states that might not be beneficial to a learner. Upon observing trained teacher DDQN policies, we noticed that they often got stuck in an endless cycle of hitting the ball back and forth in the same pattern in an effort to get the last few remaining bricks. Figure~\ref{fig:specific-game} shows a representative example from Breakout. Advanced agents are frequently in complex states, which intuitively might not be as helpful to beginning learners. Thus, for faster learning, it may initially make more sense to use samples from a a slightly weaker agent.

Other interesting results come from Robotank and Seaquest. While results suggest that using any of the selection functions results in improvements over a teacher (which may in large part be due to more frequent gradient updates), there exists a wide range in performance among the selection methods. Thus, careful tuning may be beneficial, and it would be interesting to see the effects of teaching methods when training for more environment steps.

Figure~\ref{fig:zpd} provides some intuition on the matching process for various $f_{\rm select}$ methods. The figure shows, for the same Space Invaders teacher, reward curves for the student under a representative best ahead, 5 ahead, and random ahead training run. We can see that the best ahead method uses the same ZPD teacher with reward 1007.5. The 5 ahead method starts with a ZPD teacher at reward 600 and then gradually the reward increases to the reward of the final snapshot. The ZPD reward is not monotonically increasing because it is not generally the case that reward always increases throughout the 23 saved snapshots.

It is interesting to see in Figure~\ref{fig:zpd} that the random ahead strategy experiences a more diverse reward range from its ZPD teachers. We believe that investigating this phenomenon in future work is important; can we understand and quantify the benefit one gets from randomizing a teacher?

\section{Concluding Remarks}

In this work, we have proposed an approach that merges the Zone of Proximal Development (ZPD) theory from psychology with modern deep reinforcement learning practices. We demonstrated an algorithm for determining which of several teachers should be selected to provide samples for a learner, and demonstrated results on Atari environments suggesting that it is not ideal to solely rely on the highest-reward teacher.

In this work we also raise a number of interesting questions for consideration in future work. For example, can we choose a dynamic $f_{select}$ function which does more than take a fixed number of $k$ steps ahead? One possibility would be defining a function which relies on the overlap of the state distribution of the teacher and learner. We would also like to explore how the proposed method may be used in combination with ideas from batch-constrained learning~\cite{off-policy-deeprl-2019}, or Random Ensemble Mixtures~\cite{offpolicy_atari_2019} to better train agents entirely off-policy. 

While it is only a single step towards developing complex teaching strategies, this work shows that defining teaching strategies should serve as an important step in the deep learning from demonstration pipeline, and that by choosing naive methods we are ignoring significant possible performance gains.


\medskip
\cleardoublepage 
\small
\bibliographystyle{plainnat}
\bibliography{neurips_2019}

\end{document}